\documentclass[fleqn,10pt]{wlscirep}
\usepackage[utf8]{inputenc}
\usepackage[T1]{fontenc}
\title{Generative AI and Its Impact on Personalized Intelligent Tutoring Systems}

\author[1]{Subhankar Maity}
\author[2]{Aniket Deroy}
\affil[1]{Department of Artificial Intelligence, Indian Institute of Technology Kharagpur}
\affil[2]{Department of Computer Science \& Engineering, Indian Institute of Technology Kharagpur}

\begin{abstract}
Generative Artificial Intelligence (AI) is revolutionizing educational technology by enabling highly personalized and adaptive learning environments within Intelligent Tutoring Systems (ITS). This report delves into the integration of Generative AI, particularly large language models (LLMs) like GPT-4, into ITS to enhance personalized education through dynamic content generation, real-time feedback, and adaptive learning pathways. We explore key applications such as automated question generation, customized feedback mechanisms, and interactive dialogue systems that respond to individual learner needs. The report also addresses significant challenges, including ensuring pedagogical accuracy, mitigating inherent biases in AI models, and maintaining learner engagement. Future directions highlight the potential advancements in multimodal AI integration, emotional intelligence in tutoring systems, and the ethical implications of AI-driven education. By synthesizing current research and practical implementations, this report underscores the transformative potential of Generative AI in creating more effective, equitable, and engaging educational experiences.
\end{abstract}

\begin{document}

\flushbottom
\maketitle
%
%
\thispagestyle{empty}

\section{Introduction}

Intelligent Tutoring Systems (ITS) have long been at the forefront of educational technology, aiming to provide individualized instruction and support to learners \cite{n27}. These systems are designed to offer personalized learning experiences that adapt to the unique needs and abilities of each student. Traditional ITS leverage predefined instructional materials and rule-based algorithms to adapt to learner performance \cite{n1}. However, these systems often lack the flexibility and depth necessary to cater to the diverse and dynamic needs of learners across various subjects and contexts \cite{n27}.

The advent of Generative AI, particularly large language models (LLMs) such as ChatGPT, has introduced a paradigm shift in ITS by enabling the generation of dynamic, contextually relevant educational content \cite{n2}. Generative AI algorithms can produce novel text, images, and interactive dialogues based on learned patterns from extensive datasets \cite{n28}. This capability allows ITS to create personalized learning materials that adapt to individual learners' needs in real-time \cite{n3}. For example, LLMs can generate custom questions that challenge students appropriately based on their prior responses, ensuring engagement and relevance. Additionally, they can provide context-aware feedback, guiding learners through complex problem-solving processes with explanations tailored to their unique misunderstandings \cite{n3}. Furthermore, the interactive tutoring dialogues enabled by Generative AI can simulate human-like conversations, offering instant support as students navigate their learning journeys \cite{n29}. This conversational aspect enhances engagement and fosters deeper understanding by allowing learners to explore concepts dynamically. This report explores the multifaceted applications of Generative AI within ITS, examining its potential to revolutionize personalized education. It will highlight challenges associated with its implementation—such as ensuring content accuracy and managing biases—and discuss future directions for enhancing personalized learning. By addressing these areas, we aim to provide a comprehensive understanding of how Generative AI can transform the landscape of Intelligent Tutoring Systems and improve the educational experience for learners worldwide.



\section{Generative AI in ITS: A New Paradigm for Personalized Learning}

\subsection{Automated Question Generation}

One of the most significant applications of Generative AI in ITS is automated question generation \cite{n30}. Traditional ITS rely on static question banks, which can limit the adaptability and responsiveness of the system to individual learner needs. LLMs like GPT-4 can generate a diverse array of questions tailored to the learner's current understanding and proficiency level \cite{n4}. These models can create questions that vary in difficulty, topic specificity, and cognitive demand, thereby providing a more personalized assessment of learner progress. Automated question generation also facilitates the creation of contextual and scenario-based questions that enhance learner engagement and understanding \cite{a6, a7, a8, a9, a10}. For instance, in a biology ITS, generative models can produce questions that integrate real-world applications, such as environmental impacts on ecosystems, thereby making learning more relevant and engaging \cite{n5}. Furthermore, the ability to generate an unlimited number of unique questions helps in mitigating issues related to question repetition and memorization, encouraging deeper cognitive processing and understanding.

\subsection{Personalized Feedback Mechanisms}

Effective feedback is crucial for reinforcing learning and correcting misunderstandings. Generative AI enhances ITS by providing personalized, context-specific feedback that goes beyond generic responses. LLMs can analyze a learner's input in real-time and generate detailed explanations, hints, or corrective feedback tailored to the specific errors or misconceptions identified \cite{n12}. For example, in a mathematics ITS, if a learner incorrectly applies a formula, the generative model can not only indicate the mistake but also provide a step-by-step explanation \cite{a8} to guide the learner towards the correct solution. This level of personalized feedback promotes active learning and helps learners develop a deeper understanding of the subject matter. Additionally, generative models can adapt the feedback based on the learner's progress over time, ensuring that the support provided remains aligned with their evolving needs \cite{n13}. This adaptability enhances learner engagement and motivation, as the feedback feels more relevant and supportive.

\subsection{Interactive Dialogue Systems}

Generative AI enables the creation of interactive dialogue systems within ITS, allowing for more natural and engaging interactions between the system and the learner. These dialogue systems can simulate human-like conversations, providing explanations\cite{a8}, answering questions, and guiding learners through complex concepts in a conversational manner \cite{n14}. By leveraging the contextual understanding and language generation capabilities of LLMs, ITS can offer a more immersive and interactive learning experience. Interactive dialogue systems also support scaffolding, where the system gradually increases the complexity of interactions based on the learner's proficiency. For instance, in a language learning ITS, the system can engage in increasingly complex conversations as the learner's language skills improve, providing a dynamic and responsive learning environment \cite{n15}. This interactivity not only enhances engagement but also fosters a more personalized and effective learning process.

\section{Challenges and Considerations}

\subsection{Ensuring Pedagogical Accuracy}

While Generative AI offers significant advancements in ITS, ensuring the pedagogical accuracy of generated content remains a critical challenge. LLMs are trained on vast datasets that may contain outdated, incorrect, or biased information, which can inadvertently lead to the generation of inaccurate or misleading educational content \cite{n16}. To address this, it is essential to implement robust validation mechanisms that review and verify the accuracy of AI-generated content before it is presented to learners. Hybrid systems that combine the flexibility of generative models with rule-based pedagogical guidelines can help maintain content reliability. Additionally, incorporating expert oversight and continuous monitoring can ensure that the generated content aligns with current educational standards and curricular objectives \cite{n17}. Ensuring pedagogical accuracy is paramount to maintaining the trustworthiness and effectiveness of ITS.

\subsection{Mitigating Bias and Promoting Equity}
Generative AI systems inherit biases present in their training data, which can perpetuate inequalities in educational content delivery. For example, language models may generate examples or scenarios that reflect certain cultural or socioeconomic biases, potentially disadvantaging learners from diverse backgrounds \cite{n18}. Mitigating these biases is crucial to promoting equity and inclusivity in education. Strategies for bias mitigation include using more diverse and representative datasets, implementing fairness algorithms, and conducting regular audits of AI-generated content to identify and address biased outputs \cite{n19}. Additionally, involving educators and stakeholders from diverse backgrounds in the development and evaluation process can help ensure that the ITS caters to a broad range of learners and perspectives. Promoting equity in ITS not only enhances the learning experience for all students but also aligns with broader educational goals of fairness and inclusivity.

\subsection{Maintaining Learner Engagement}

While Generative AI can enhance personalization and interactivity, maintaining learner engagement over time remains a challenge. Over-reliance on AI-generated content without sufficient human oversight can lead to disengagement if the content lacks depth or fails to address the nuanced needs of learners \cite{n20}. Balancing AI-driven personalization with meaningful educational experiences is essential to sustaining learner motivation and interest. Incorporating elements such as gamification, adaptive learning pathways, and varied instructional strategies can help maintain engagement. Additionally, ensuring that the AI-generated interactions are contextually relevant and aligned with the learner's goals can enhance the perceived value and relevance of the ITS \cite{n21,n22, n23}. Engaged learners are more likely to achieve better learning outcomes, making engagement a critical factor in the success of ITS.

\section{Future Directions}

\subsection{Multimodal AI Integration}
Future advancements in ITS will likely involve the integration of multimodal AI, combining text, speech, and visual inputs to create richer and more immersive learning experiences. Multimodal models can process and generate content across different formats, enabling ITS to offer interactive simulations, visual explanations, and spoken dialogues that cater to various learning styles \cite{n24}. This holistic approach can enhance comprehension and retention by engaging multiple senses and providing diverse modes of information delivery.

\subsection{Emotional Intelligence and Affective Computing}

Incorporating emotional intelligence into ITS can further personalize and enhance the learning experience. Affective computing enables ITS to recognize and respond to learners' emotional states, such as frustration, confusion, or excitement, thereby providing more empathetic and supportive interactions \cite{n25}. By adapting their responses based on the learner's emotional cues, ITS can create a more supportive and motivating learning environment, fostering resilience and persistence in learners.

\subsection{Ethical Implications and Policy Development}

As Generative AI becomes more prevalent in education, addressing ethical implications is paramount. Issues such as data privacy, informed consent, and the responsible use of AI must be carefully considered \cite{n26}. Developing comprehensive policies and guidelines that govern the deployment and use of AI in educational settings is essential to ensure that these technologies are used ethically and responsibly. Collaboration between educators, policymakers, technologists, and ethicists will be crucial in shaping the future landscape of AI-driven education.

\section{Conclusion}

Generative AI holds transformative potential for Intelligent Tutoring Systems, offering unprecedented levels of personalization, adaptability, and interactivity in education. By leveraging large language models, ITS can generate dynamic educational content, provide nuanced feedback, and engage learners through interactive dialogues tailored to individual needs. However, realizing this potential requires addressing significant challenges, including ensuring pedagogical accuracy, mitigating biases, and maintaining learner engagement. Future advancements in multimodal AI, emotional intelligence, and ethical policy development will further enhance the effectiveness and equity of AI-driven education. As researchers and practitioners continue to innovate and collaborate, Generative AI is poised to redefine personalized learning, making education more responsive, accessible, and effective for diverse learners worldwide.

\bibliography{sample}

\end{document}